\documentclass[10pt,twocolumn,letterpaper]{article}

\usepackage[pagenumbers]{cvpr} % To force page numbers, e.g. for an arXiv version

\usepackage{graphicx}
\usepackage{amsmath}
\usepackage{amssymb}
\usepackage{booktabs}
\usepackage{algorithm}
\usepackage{algorithmic}

\usepackage[pagebackref,breaklinks,colorlinks]{hyperref}

\usepackage[capitalize]{cleveref}
\crefname{section}{Sec.}{Secs.}
\Crefname{section}{Section}{Sections}
\Crefname{table}{Table}{Tables}
\crefname{table}{Tab.}{Tabs.}

\def\cvprPaperID{*****} % *** Enter the CVPR Paper ID here
\def\confName{CVPR}
\def\confYear{2022}

\begin{document}

\title{Learning Sparse Latent Representations for Generator Model\thanks{Work In Progress}
}

\author{Hanao Li\\
Stevens Institute of Technology \\
hli136@stevens.edu\\
\and
Tian Han\\
Stevens Institute of Technology\\
than6@stevens.edu
}

\maketitle

%%%%%%%%% ABSTRACT
\begin{abstract}
   Sparsity is a desirable attribute. It can lead to more efficient and more effective representations compared to the dense model. Meanwhile, learning sparse latent representations has been a challenging problem in the field of computer vision and machine learning due to its complexity. In this paper, we present a new unsupervised learning method to enforce sparsity on the latent space for the generator model with a gradually sparsified spike and slab distribution as our prior. Our model consists of only one top-down generator network that maps the latent variable to the observed data. Latent variables can be inferred following generator posterior direction using non-persistent gradient based method. Spike and Slab regularization in the inference step can push non-informative latent dimensions towards zero to induce sparsity. Extensive experiments show the model can preserve majority of the information from original images with sparse representations while demonstrating improved results compared to other existing methods. We observe that our model can learn disentangled semantics and increase explainability of the latent codes while boosting the robustness in the task of classification and denoising.
\end{abstract}

%%%%%%%%% BODY TEXT
\section{Introduction}

Over the recent years, there are many approaches that study the mapping between the high dimensional observations and low dimensional latent representations. Some of the most influential works and their variants include Variantional Auto-Encoders (VAE) \cite{kingma2013auto,rezende2014stochastic,rezende2015variational} and Generative Adversarial Networks (GAN) \cite{goodfellow2014generative,radford2015unsupervised,karras2019style}. Although they have shown impressive results, these works emphasis  on learning dense and distributed latent representations rather than disentangled or sparse latent codes. Therefore, individual latent codes are essentially a black box to us and they can be hard to explain. Moreover, the dense models are in general not robust to noise. Their redundant representations are sensitive to perturbations since small changes on the latent code will lead to different outputs via nonlinear generator network. On the contrary, sparse representations suffer less from these issues. They are more robust to noises and will lead to a better performance. They also offer other great advantages compared to dense representations. The sparse representations can provide more explainability on the latent space. They carry out semantic information and encode them into a small subset of latent dimensions to make the model more interpretable. 

Many of the existing works have shown great performance on learning disentangled or explainable latent variables \cite{DBLP:conf/icml/ReedSZL14,DBLP:conf/iclr/PlumeraultBH20}. These works aim to learn a disentangled representation in a supervised way by incorporating the extra factor information. \cite{DBLP:conf/nips/ChenCDHSSA16,DBLP:conf/ijcnn/ChrysosKYA21,DBLP:journals/corr/abs-1906-06034} uses GAN to learn disentangled semantics in unsupervised setting but they can not infer the latent variables without adding extra encoder model or optimization step. In addition, \cite{mathieu2019disentangling} imposes the decomposition using a desired structure on the vanilla VAE model to learn disentangled latent representations. \cite{liu2020towards} adds gradient-based attention to the VAE model to learn improved latent space disentanglement. Beta-VAE incorporates the extra adjustable parameter $\beta$ to sacrifice the reconstruction ability while obtaining an more interpretable factorised latent representation \cite{higgins2016beta}. These approaches mainly focused on learning disentangled latent representations with isotropic Gaussian as their prior distribution. Meanwhile, their learned latent representations are not parsimony and they are not as robust as sparse representations. Hence, we are motivated to learn a model that can learn a sparse representation as well as disentangled latent semantics. 

In this paper, we present a new learning method for learning sparse representations using gradually sparsified spike and slab distribution as our prior belief. We use maximum likelihood estimation to train the model and Monte Carlo Markov Chain sampling \cite{neal2011mcmc} to infer the latent variables to solve the intractable expectation. In practice, we disable the random noise in the Langevin Dynamics sampling and the sampling process essentially becomes a gradient based inference. We show theoretically latent codes that do not capture important information will be forced towards zero under the spike and slab prior regularization in the inference step. The learned sparse latent codes can then be used to update the generator model under MLE. During training, the process of using zero initialization with gradient based inference to infer the latent variables shares the similar spirit as \cite{nijkamp2020learning} in their proposed model. With the above single network learning scheme, we can perform exact inference in the generator model without the need of designing an extra encoder network for approximated inference. The learned latent representations can be more accurate and effective.

\vskip -0.3in

\iffalse
Our contributions are summarized as below:
\begin{itemize}
  \item  We present a new learning method to learn a sparse latent representations via gradually sparsified spike and slab prior to avoid dead latent codes. 
  \item  We propose to use iterative penalized gradient-based inference in a short-run manner to learn informative sparse latent variables.
  \item We conduct extensive experiments to demonstrate our sparse model can capture essential information of the original observation and lead to interpretable as well as robust representation. 
  %'s promising results in terms of reconstruction ability and image quality compared to existing methods. It also demonstrates improved explainability and boosted robustness in the task of classification and denoising.
  %\item We conduct extensive experiments to show our method's promising results in terms of reconstruction ability and image quality compared to existing methods. It also demonstrates improved explainability and boosted robustness in the task of classification and denoising.
\end{itemize}
\fi

\section{Related Works}

\subsection{Learning Linear Sparse Representations}

There have been many applications and algorithms that involve the learning of sparse representations \cite{wright2008robust,yang2010image,yang2011robust}. One of the most fundamental methods to learn a sparse representation is sparse coding \cite{lee2007efficient,olshausen1997sparse}. It is an approximation strategy that is developed to solve the optimization problem of finding optimal weighted linear combinations of the basis matrix and coefficient matrix from an over-complete dictionary.

The goal of sparse coding aims to learn a meaningful sparse representation without losing much details while keeping only a small set of latent codes to have strong activation. \cite{zeiler2010deconvolutional} proposes Convolutional Sparse Coding (CSC) to learn rich features and it is based on convolutional decomposition of images under sparsity constraint. \cite{bristow2013fast} produces a fast convolutional sparse coding algorithm with super linear convergence. \cite{evtimova2021sparse} proposes using iterative shrinkage-thresholding algorithm to directly regularize the latent variables and learn sparse coding. These works have focused on learning linear mapping of sparse coding, it sometimes does not have the flexibility to learn complex mapping. Especially in the era of neural networks, non-linear relationships are more expressive and preferred nowadays.

\subsection{Learning Non-Linear Latent Representations}

Learning mapping between the observation and the latent space has also gained increasing popularity recently \cite{kingma2013auto,goodfellow2014generative}. Alternating back-propagation (ABP) is an algorithm that uses a unified probabilistic framework to train the generator network \cite{DBLP:conf/aaai/HanLZW17}. The generator uses a deep generative network to obtain a complex non-linear mapping from the latent space to the observation. It adopts persistent chain for MCMC sampling \cite{neal2011mcmc}, i.e., warm start sampling, throughout the training to obtain their posterior latent samples. \cite{nijkamp2020learning} proposes using short-run non-persistent chain when sampling the posterior. However, these models have dense latent representations and they do not aim to learn a sparse representation under their training procedure. The entangled latent variables have minimal interpretability and they are not robust to noisy data. \cite{han2018learning} extends the vanilla ABP for disentangled latent representations learning, but the learned latent variables are not sparse.  \cite{xing2020inducing} proposes a hierarchical AND-OR generator model to learn meaningful interpretations of the latent variables. 

Currently, there are limited works that learn a sparse representation on the latent space. One of the most notable and relevant example is Variational Sparse Coding (VSC) \cite{tonolini2020variational}. It builds upon variational inference method to learn the non-linear mapping between observation and the latent space. VSC presents using the scaled sigmoid step function to approximate the behavior of Dirac Delta function. However, due to the nature of variational models, VSC needs to do approximations for the true posterior of the latent variable instead of doing exact inference. It also needs to carefully design an extra inference network that learns the mapping from the observation to the latent space and their learned latent representations are vulnerable to noisy data. The modified KL divergence regularization in VSC also can only approximate the target distribution without learning much semantics. In contrast, our gradient based learning and inference utilizes the information from top-down generator that can make our latent representations more accurate and effective. It can also push non-informative latent towards zero so that activated dimensions can contain semantic information.

\section{Proposed Method}

\subsection{Spike and Slab Prior}
We present a new learning method to learn sparse representations using spike and slab distribution as our prior belief for the latent variables on the generator model. The distribution consists of spike variable and continuous slab variable \cite{mitchell1988bayesian,goodfellow2012large}. The spike variable has a probability $\alpha_1$ that determines whether the latent variables take values from the standard Gaussian or the slab variable distribution where it can be either a Dirac delta function or another Gaussian distribution. Since the Dirac delta function is not differentiable, we instead adopt a Gaussian distribution centered at 0 with a small variance to approximate the behavior of the Dirac delta function. The prior distribution can still be regarded as spike and slab but now it can also be viewed as a Gaussian mixture model with weights $\alpha_1 + \alpha_2 = 1$ as shown in Equation \ref{e1}. 
\begin{equation}
    z \sim p_{ss}(z) = \alpha_1 N(0, \sigma_{1}^2) + \alpha_2 N(0, \sigma_{2}^2)
    \label{e1}
\end{equation}
where $\sigma_1^2$ is fixed to be 1 as the variance of standard Gaussian distribution and $\sigma_2^2$ is the variance of slab Gaussian distribution.
With this prior, the sparsity of our latent variable $z$ can be determined by changing the value of $\alpha_1$. Theoretically, we should have a small $\alpha_1$ and $\sigma_2^2$ to induce sparsity. With a small $\alpha_1$, $\alpha_2$ will be large and it's more likely to sample points from the slab variable distribution. When we have a small $\sigma_2^2$ simultaneously, most points sampled from this distribution $p_{ss}(z)$ will be small and close to 0. Thus we can induce a sparse representation on the latent space in this setting.
\subsection{Latent Inference and Model Learning}
For each given observation $x$, we assume there is a corresponding latent variable $z$. The generator model $G$ that maps the latent variables into observations can be represented as:
\begin{equation}
\label{e2}
    x = G_{\theta}(z) + \epsilon; \, \, \epsilon \sim N(0, \sigma^2 I_D)
\end{equation}
where $x \in \mathbb{R}^d$, $\sigma$ is the pre-specified standard deviation of the noise vector $\epsilon$ and the generator $G$ is parameterized by a top-down neural network with weights $\theta$. Equation \ref{e2} implies that the conditional distribution $p_{\theta}(x|z)$ is also a Gaussian distribution with $p_{\theta}(x|z) \sim N(G_{\theta}(z), \sigma^2 I_D)$.
Given these two distributions $p_{ss}(z)$ and $p_{\theta}(x|z)$ , our complete-data log-likelihood between observation $x$ and latent variable $z$ can be formulated as follows:
\begin{equation}
\begin{aligned}
\label{e3}
 \log  p_{\theta}(x, z) & = \log [p_{ss}(z)p_{\theta}(x|z)]  \\ 
& =-\frac{|x - G_{\theta}(z)|^2}{2\sigma^2} + \log p_{ss}(z) + \log \frac{1}{\sqrt{2\pi}\sigma}
\end{aligned}
\end{equation}
The log of joint distribution consists of the reconstruction error and log prior penalty term. This model can be trained using maximum likelihood estimation. The observed-data log-likelihood $L(\theta)$ given observations $\{x_1, x_2, ..., x_n\}$ can be written as:
\begin{equation}
L(\theta) = \sum_{i=1}^n \log p_{\theta}(x) = \sum_{i=1}^n \log  \int_z p_{\theta}(x_i, z_i)dz 
\end{equation}
Differentiate with respect to weights $\theta$, the gradient of log-likelihood can be derived as:
%with respect to the weights $\theta$ can be calculated into:
\begin{equation}
\label{e5}
\frac{\partial}{\partial \theta} L(\theta) = \sum_{i=1}^n E_{p_{\theta}(z_i|x_i)}[\frac{\partial}{\partial \theta} \log p_{\theta}(x_i,z_i)]
\end{equation}
However this posterior inside the expectation can be intractable to sample from, we therefore consider using non-persistent short-run Langevin Dynamics with spike and slab prior to obtain a sparse latent representation. Then the generator posterior distribution of $z$ in Equation \ref{e5} can be updated via:
\begin{equation}
\label{e6}
     z_{\tau+1} = z_{\tau} - \frac{s^2}{2}\frac{\partial}{\partial z}[\frac{|x - G_{\theta}(z)|^2}{2\sigma^2} - \log p_{ss}(z)] + s\epsilon_{LD, \tau} 
\end{equation}
where $s$ denotes the learning step size, $\epsilon_{LD}$ denotes noise diffusion term and $\tau$ is the time step of Langevin Dynamics. As $\tau \rightarrow \infty$, the posterior distribution of $z$ will converge to $p_{\theta}(z|x)$ given a small step size $s$. This enables us to initialize $z_0$ from any fixed distribution $p_0$. The starting distribution can be either Gaussian, spike and slab or zero initialization. 

In practice, the noise diffusion term in the Langevin Dynamics can be disabled, so the posterior inference of $z$ becomes the gradient-based inference that is driven by reconstruction error and the latent log-prior penalty. The whole process leads to maximum a posteriori (MAP) and is performed by penalized gradient descent. We observe the gradient-based inference with zero initialization of $z_0$ leads to slightly better performance since the added noise term will distort the learned sparse structure, we thus adopt such scheme for our experiments since they are more suitable with the goal of achieving sparsity.
%and use zero initialization to get rid of the randomness from the probability distribution. The learning of latent variable has essentially become a penalized gradient descent. 

In Equation \ref{e6}, the derivative of log of spike and slab prior regularization term with respect to the latent variable $z$ can be represented as:
\begin{equation}
\label{e7}
    \frac{\partial}{\partial z}\log p_{ss}(z) = -\frac{z}{\sigma_1^2} + \frac{R_1 R_2 z}{e^{-R^2 \frac{z^2}{2}}+R_1}
\end{equation}
where we denote $R_1 = \frac{(1-\alpha_1) \sigma_1}{\alpha_1 \sigma_2}$ and $R_2 = \frac{\sigma_2^2 - \sigma_1^2}{\sigma_1^2 \sigma_2^2}$. See Appendix A.1 for detailed derivation. This regularization term can push the latent codes towards zero by giving it a larger gradient if it does not capture meaningful information. When we enforce a highly sparse latent power (e.g. $\alpha_1 = 0.01$) at the start of training, some of the latent codes will not have the opportunity to learn anything before get pushed towards zero. We demonstrate by slowing decay the value of $\alpha_1$ in the course of training, individual latent variables can gradually capture semantic knowledge. This can help alleviating the problem of dead latent codes.

After latent codes have been inferred, they can be plugged in Equation \ref{e3} to form the complete-data log-likelihood. The generator can be learned using stochastic gradient decent:
%we obtain the posterior of latent variables from Langevin Dynamics sampling, our model parameters can then be updated using the complete-data log-likelihood together with the observed $x_i$. By taking the gradient in Equation \ref{e3} with respect to the parameters $\theta$, we update the model using stochastic gradient decent:

\begin{equation}
\label{e8}
\theta_{t+1} = \theta_{t} + \eta\frac{1}{n}\sum_{i=1}^n E_{p_{\theta}}(z_i|x_i)[\frac{\partial}{\partial \theta} \log p_{\theta}(x_i, z_i)]	
\end{equation}

%\begin{equation}
%\label{e8}
%\frac{\partial}{\partial \theta}  L(\theta) = \sum_{i=1}^n  \frac{x_i - %G_{\theta}(z_i)}{\sigma^2}\frac{\partial}{\partial \theta}G_{\theta}(z_i)	
%\end{equation}

\subsection{Theoretical Understanding}
From Equation \ref{e1}, we observe that the spike and slab distribution is essentially a Gaussian mixture model with two Gaussian components when the spike variable uses Gaussian distribution to approximate the Dirac Delta function. We denote $p(C_i) = \alpha_i$ as the prior probability of component $i$. The posterior probability of the component $p(C_i|z)$ given latent variable is bounded between 0 and 1, and can be represented as follows:
\begin{equation}
\begin{aligned}
   p(C_i|z) &= \frac{\alpha_i N(0, \sigma_i^2)}{\alpha_i N(0, \sigma_i^2) + \alpha_j N(0, \sigma_j^2)}  \\
 & = \frac{\frac{\alpha_i}{\sigma_i}}{\frac{\alpha_i}{\sigma_i}+\frac{\alpha_j}{\sigma_j}e^{z^2(\frac{1}{2\sigma_i^2}-\frac{1}{2\sigma_j^2})}}; \, \, \, i, j \in \{1, 2\}; i \neq j
\end{aligned}
\end{equation}
See Appendix A.2 for detailed derivation. Theoretically, one of the component's variance must be small to induce sparsity, so we let $\sigma_2^2 < \sigma_1^2 = 1$ and $\frac{1}{2\sigma_1^2} - \frac{1}{2\sigma_2^2}$ will always be negative. With these premises, we can derive and rewrite the gradient of the logarithm of the spike and slab prior from Equation \ref{e7} in terms of the posterior probability $p(C_i|z)$ as below:

\begin{equation}
\label{e9}
\begin{aligned}
 \frac{\partial}{\partial z}  \log p_{ss}(z) 
 &  = -p(C_1|z)\frac{z}{\sigma_1^2}-p(C_2|z)\frac{z}{\sigma_2^2}
\end{aligned}
\end{equation}
See Appendix A.3 for detailed derivation. Since we want to have more points sampled from the Gaussian distribution with small variance, we set $\alpha_1 < \alpha_2$ and then $p(C_2) > p(C_1)$. 

When latent variable $z$ tends to be large, $p(C_1|z)$ will approach to 1 while $p(C_2|z) * z$ will approach to 0, so the gradient term will become $-\frac{z}{\sigma_1^2}$ since it is more likely to be sampled from standard Gaussian component. But when $z$ is relatively small, $p(C_1|z)$ will approach to 0 while $p(C_2|z)$ will approach to 1, and the gradient will become $-\frac{z}{\sigma_2^2}$. Since $\sigma_2^2$ is a very small number, we will have a larger gradient for small value of $z$ which means a stronger power to push it towards 0. 

This indicates that the value of $z$ for this dimension must contain important information to overcome this power. Otherwise, if $z$ holds inconsequential information,  the prior term will penalize it and thus enforce sparsity to a certain degree.

\begin{algorithm}[ht]
   \caption{Sparse Latent Representations Learning}
   \label{alg}
\begin{algorithmic}
   \STATE {\bfseries Input:} observations $x_i$, langevin steps $l$, number of epochs $T$, learning rate $\eta$, current sparsity level $\alpha_C$, sparsity decay constant $\gamma$, sparsity threshold $\alpha_T$.
   \STATE Let $t = 0$;
   \REPEAT
   \STATE \textbf{Sparsity Decay}: If $\alpha_C > \alpha_T$, then $\alpha_C = \alpha_C - \gamma$.
   \STATE \textbf{Latent Initialization}: For each $x_{i}$, initialize a new $z_i$ from fixed prior distribution $z_i \sim p_0(z)$. 
   \STATE \textbf{Latent Update}: Infer the maximum of posterior of $z_i$ using gradient-based inference $l$ steps according to Equation \ref{e6} and \ref{e7}.
   \STATE \textbf{Model Update}: Fix the inferred $z_i$ and observation $x_i$,  update the model parameters with learning rate $\eta$ according to Equation \ref{e8}.
   \STATE Let $t = t + 1$;
   \UNTIL{$t = T$}
\end{algorithmic}
\end{algorithm}

\section{Experiments}

We test our model on MNIST \cite{lecun2010mnist}, Fashion-MNIST \cite{xiao2017fashion}, CelebA \cite{liu2015deep} and SVHN \cite{netzer2011reading} datasets. MNIST and Fashion-MNIST are grey scaled images while CelebA and SVHN are color images. CelebA images are cropped and resized to dimensions of 64 by 64.

\noindent \textbf{Experimental Settings:} All training and testing images are scaled to [0, 1]. Unless otherwise stated, we use 200 latent dimensions for MNIST and Fashion-MNIST, 400 for SVHN and 800 for CelebA. We fix batch size = 100, $\sigma$ = 0.3, langevin step size = 0.1, langevin steps = 30, slab variable variance = 0.1, and $\alpha_1 = 0.01$. Initial sparsity is set to 1 and the decay constant $\gamma$ is 0.033. We utilize cold start to train the model with zero initialization at the start of every epoch. Adam optimizer \cite{kingma2014adam} is used with a learning rate of 0.0001. 

For a fair comparison with the VSC model \cite{tonolini2020variational}, we adopt their model structure which consists of 1 hidden layer with 400 hidden units followed by ReLU activation and sigmoid non-linearity as the output layer for MNIST and Fashion-MNIST, and we use 2 hidden layers with 2000 hidden units for CelebA and SVHN.

\subsection{Sparsity Control}

We first verify our training with gradient-based inference can enforce sparsity on the latent codes. In this experiment, we demonstrate our model can learn a sparse representation of the latent codes effectively by simply changing the prior probability $\alpha_1$. We can obtain the learned latent representations of the current observation using gradient based inference. The value of $\alpha_1$ is set to be [1, 0.5, 0.01] for comparison and illustration. 

\begin{figure}[ht]
\vskip 0in
\begin{center}
\centerline{\includegraphics[width=\columnwidth]{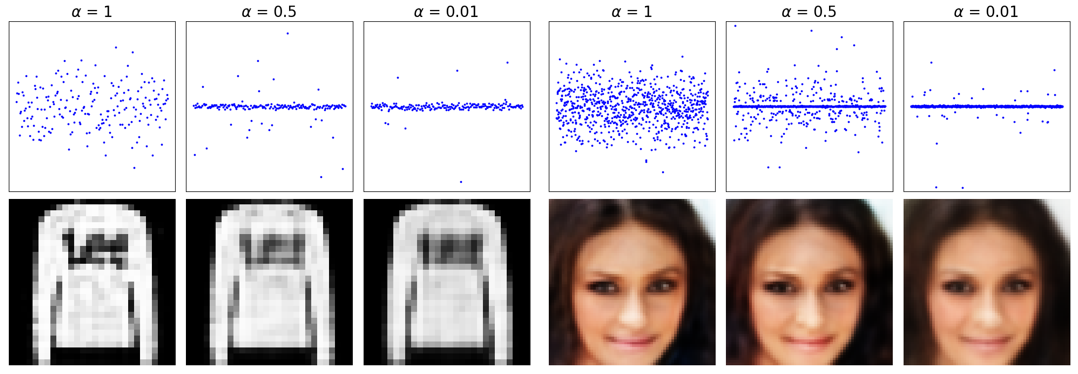}}
\caption{Latent Representations with Different Levels of sparsity level $\alpha_1$. \textbf{Top:} Learned Latent Representations. \textbf{Bottom:} Corresponding  Reconstructed Images.}
\label{sc}
\end{center}
\end{figure}

We show in Figure \ref{sc} that when the prior probability for standard Gaussian component $\alpha_1 = 1$, learned latent representation is dense and nearly all the latent codes are activated. When we gradually decrease the value of $\alpha_1$ to 0.01, there are fewer latent codes with stronger activation and the rest of them have been pushed towards 0. Thus we have shown our algorithm is effective and has learned the sparse representations. Meanwhile, we can also observe the reconstructed images with sparse latent codes do not lose much information compared to dense latent codes on given observation. 
%Our model has shown capability to learn a sparse representation on the latent space.

\subsection{Reconstruction of Sparse Representations}

The ability to reconstruct is one of the important factor to measure whether the learned latent codes can capture the essential information of the observations. We show that our model can reconstruct the testing images well with only a small number of activated latent variables on various datasets. We compare our results with VSC using same value of $\alpha$ ($\alpha=0.01$) \cite{tonolini2020variational}, VAE \cite{kingma2013auto}, Beta-VAE ($\beta = 4$) \cite{higgins2016beta} and short-run inference model \cite{nijkamp2020learning}. We assess the performance of reconstruction ability by comparing the reconstruction quality and evaluating peak signal-to-noise ratio.

\begin{figure}[ht]
\begin{center}
\includegraphics[width=\linewidth]{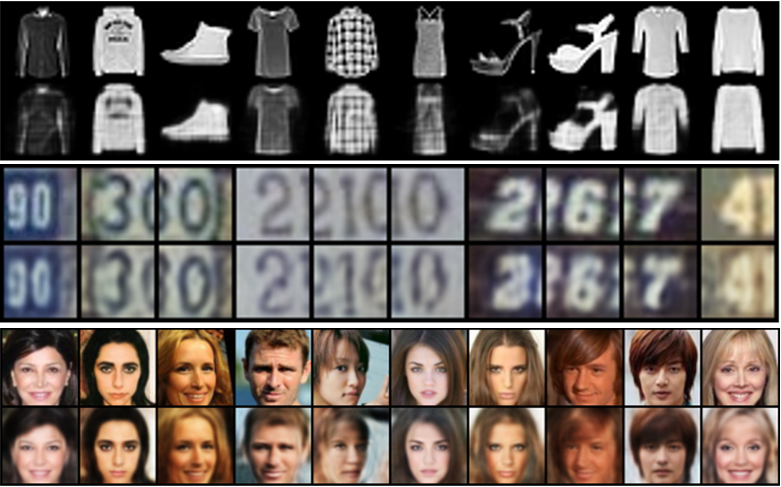}
\caption{Reconstructed Images on various datasets. \textbf{Top:} Original Images. \textbf{Bottom:} Reconstructed Images.}
\label{mse}
\end{center}
\end{figure}

\begin{table}[ht]
\centering
\begin{tabular}{lllll}
\hline\noalign{\smallskip}
Model & MNIST & Fashion & CelebA & SVHN\\
\noalign{\smallskip}
\hline
\noalign{\smallskip}
VSC  & 32.73 & 34.45 & 40.11 & 41.31\\
VAE & 38.71 & 38.69 & 47.74 & 47.53\\
Beta-VAE & 33.68 & 33.99 & 40.82 & 39.64\\
Short-run  & 39.91 & 43.61 & 53.56 & 58.41\\
Ours ($\alpha = 0.01$) & 37.20 & 39.02 & 49.37 & 50.17
\\
\hline
\end{tabular}
\caption{Peak Signal-to-Noise Ratio (dB). Higher PSNR metric means better reconstruction quality.}
\label{mset}
\end{table}

\begin{figure}[ht]
\begin{center}
\includegraphics[width=\linewidth]{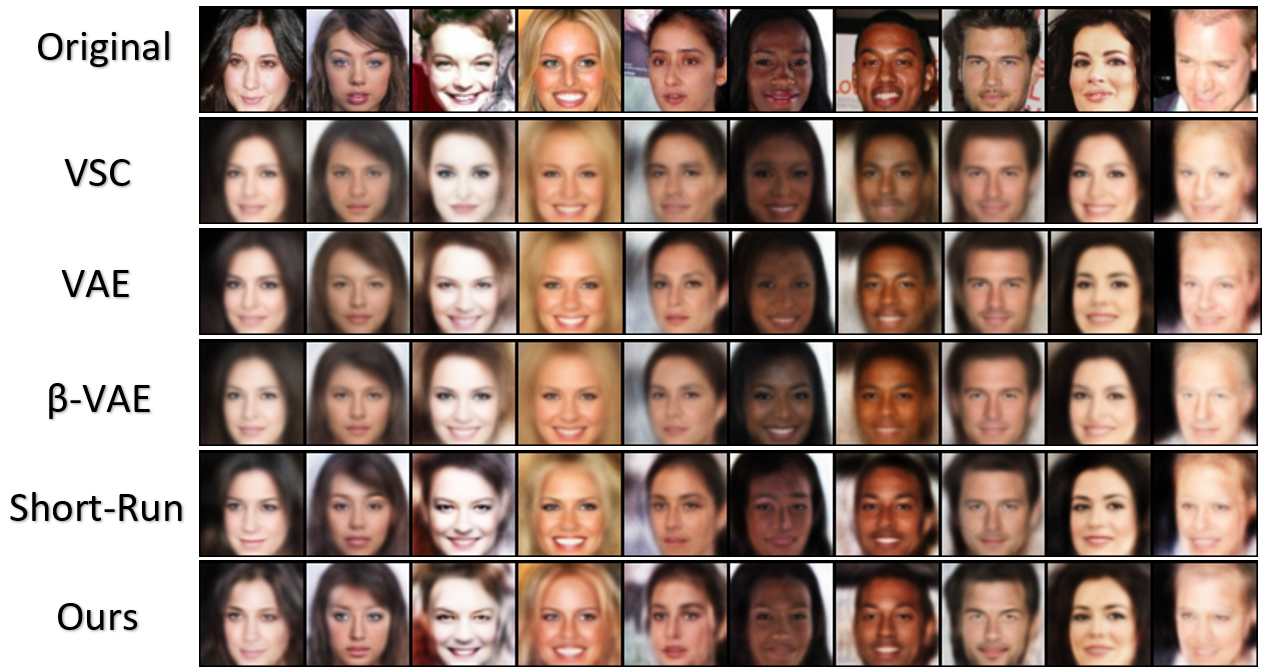}
\caption{Reconstructed Face Images with various models. \textbf{First Row} Original Images. \textbf{Second the Last Rows:} Comparisons of reconstructed images.}
\label{com}
\end{center}
\end{figure}

In Table \ref{mset}, our model can outperform both VSC and $\beta$-VAE while being competitive to the dense VAE model. Although our model obtained lower PSNR than short-run model, we have only used very few activated latent codes.  As demonstrated in Figure \ref{mse} and \ref{com}, the reconstructed images using learned sparse latent representations can preserve majority of the information and learn more accurate semantics such as smile and hairstyle than other models.

\subsection{Latent Code Exploration}

\begin{figure}[ht]
\begin{center}
\centerline{\includegraphics[width=\columnwidth]{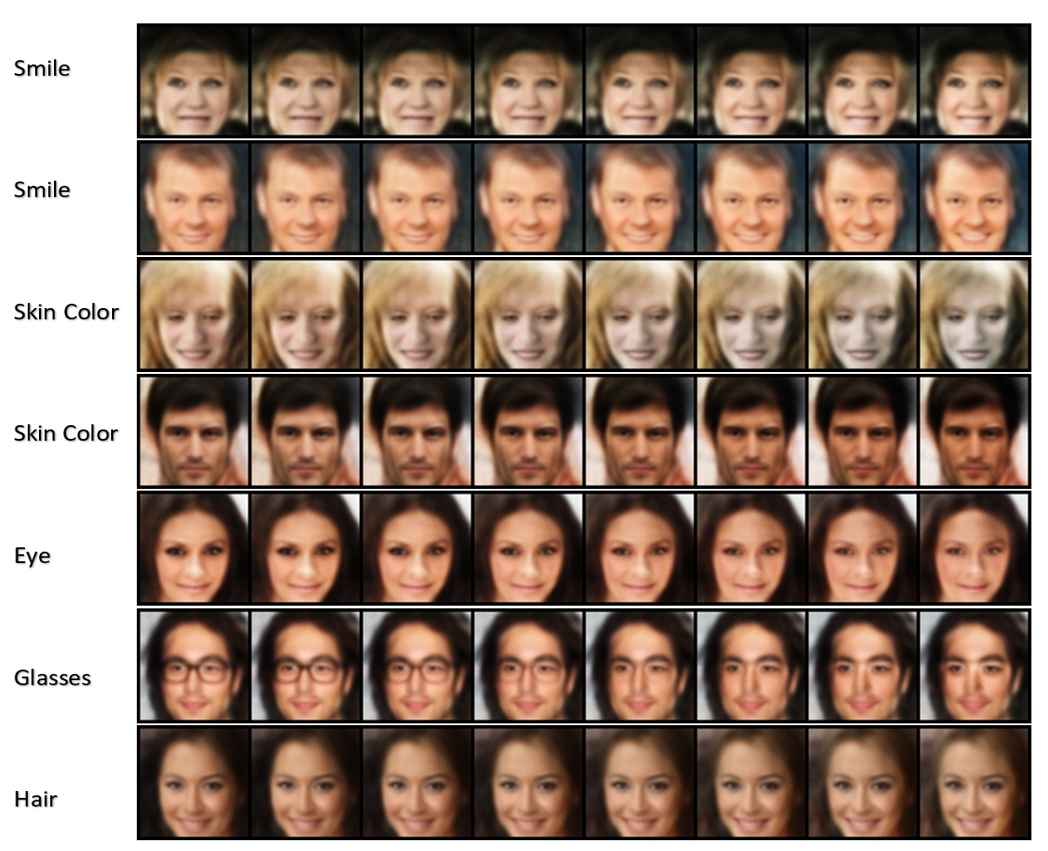}}
\caption{Latent Traversal using CelebA dataset. First Column: Reconstructed image given observation. Second to Last Columns: Reconstructed Images after incrementally altering single activated latent code}
\label{traversal}
\end{center}
\end{figure}

With a well-learned model, a meaningful sparse latent representation can be captured. In this experiment, we show that our model can learn semantic meanings from the images and encode them into individual latent codes. By changing one of the activated dimension in the latent space, we can see disentangled changes in the reconstructed image, and thus the model makes the latent codes more explainable. In Figure \ref{traversal}, we demonstrate by altering single latent dimension, changes of disentangled factors can be observed while keeping other features unaffected. The learned latent codes have shown their capacity to discover disentangled factors such as facial features (with or without glasses) and expressions (smile or non-smile) from the training dataset in an unsupervised manner.

Additionally, we train a Fashion-MNIST dataset with 30 latent dimensions to examine features learned with short-run and our model. Fashion-MNIST contains 10 classes and classes from the same category such as Top and Pullover share common structures. Explainable models should be able to extract such information and encode them into corresponding latent representations. For all testing data, we separate them by their class label and their latent representations can be obtained using Langevin Dynamics or gradient based inference. The activation threshold is set to be 0.2. Activated latent codes are set to 1 and non-activated codes are set to 0. Then we average all the latent codes by class and plot the heatmap to examine the features learned from the dataset. 

\begin{figure}[ht]
  \begin{minipage}{0.50\linewidth}
    \centering
    \includegraphics[width=\linewidth]{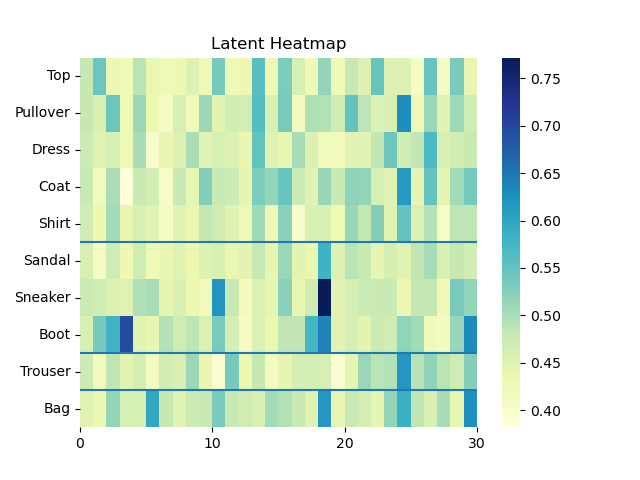}
    %\rule{6cm}{6cm} %to simulate an actual figure
  \end{minipage}%
  \begin{minipage}{0.50\linewidth}
    \centering
    \includegraphics[width=\linewidth]{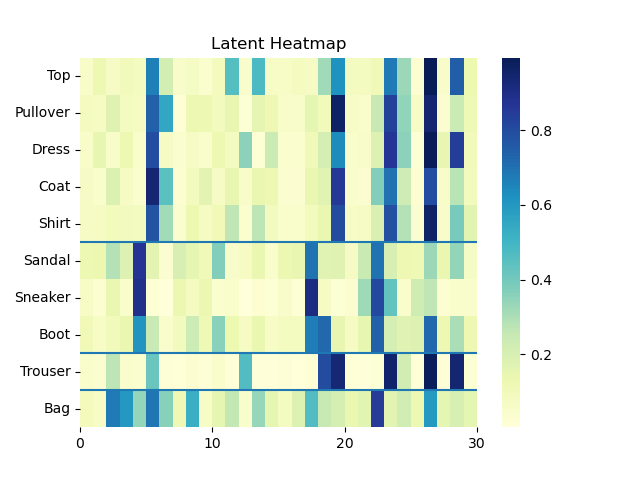}
    
    \end{minipage}
\caption{Heatmap of activated latent codes. \textbf{Left:} Short-run model. \textbf{Right:} Our model. Different categories are separated by the horizontal line. First five rows contain upper wear, middle three rows consist of foot wear and the last twos row are bottom and hand wear. Objects with the same category should have similar activated latent maps with little variations.}
\label{heat}
\end{figure}

For short-run model on left side of Figure \ref{heat}, we observe learned latent codes do not form a specific structure. Most of their latent representations are evenly separated among all 30 latent dimensions. There aren't many latent variables with significant activation. Meanwhile, on the right side of Figure \ref{heat}, we can clearly visualize the learned latent structure for each Fashion class. Each category forms an apparent pattern with only few  activated codes while most of the latent dimensions are kept inactivated. These activated codes have strong activation and some other latent codes do not activate for certain categories. This can make the learned latent variables more interpretable when altering their latent code.

\subsection{Latent Code Classifier}
We examine whether our learned sparse representations can retain important information and increase the robustness of the classifier. We test the model with different number of latent dimensions varying from 10 to 250. We randomly select 3000 images from the MNIST training dataset and encode the training observations to the latent codes as input to train a one hidden linear layer classifier followed by ReLU activation. Then, we encode testing dataset to the latent space to test the accuracy of encoded latent codes using trained classifier. 

\begin{figure}[ht]
\begin{center}
\centerline{\includegraphics[width=\columnwidth]{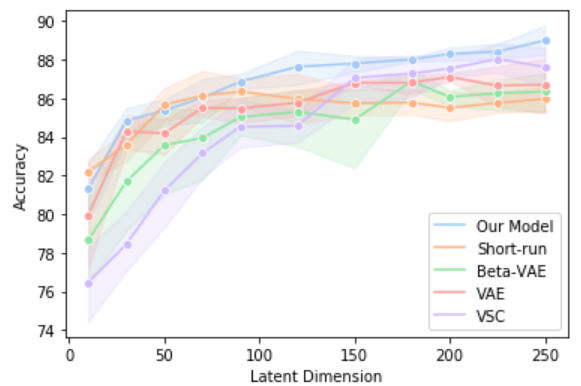}}
\caption{Classification Accuracy on MNIST datasets using encoded latent variables. The accuracy results are averaged over 5 trials. Our model obtains higher accuracy with increased latent dimensions. Dense models will saturate and lose accuracy over time. }
\label{clas}
\end{center}
\end{figure}

We see the classification result in Figure \ref{clas}, our model can encode key class information into the latent variables without losing accuracy. It can also outperforms other models at different latent dimensions. When the dimension of the latent space is increased, sparse models can fill in more details to further help with the distinguish of digits. In contrast, dense models will saturate and the accuracy will stuck at some point. Therefore, we show that our learned sparse representations can lead to a more robust classifier. The classifier learned with sparse codes will not overfit as the dimension increases. 

\begin{figure}[ht]
\begin{center}
\centerline{\includegraphics[width=\columnwidth]{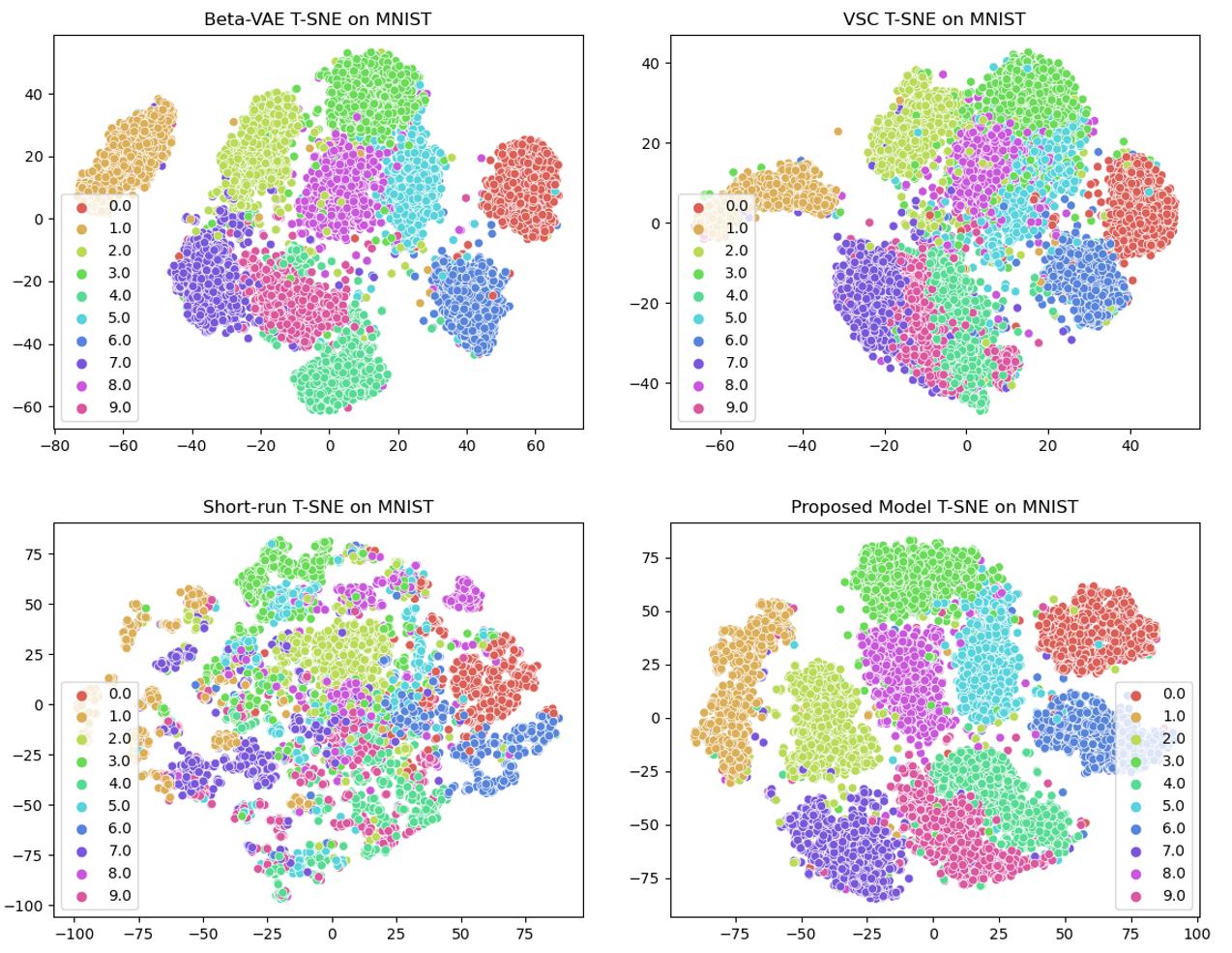}}
\caption{T-SNE plots of different models. Top left: Beta VAE; Top Right: VSC; Bottom Left: Short-run model; Bottom Right: Proposed Model.}
\label{tsne}
\end{center}
\end{figure}

We further show t-SNE plots using learned latent codes with 200 latent dimensions for MNIST dataset to verify our classifier results. In Figure \ref{tsne}, the sparse latent representations learned by our model can be well separated into different clusters compared to existing models. The ability to separate the clusters implies our model have captured pivotal knowledge of each class which leads to a more robust classifier.

\subsection{Robust Latent Representations}

A sparse latent representations should also be robust to noisy images semantically and visually as it can encode important structural information from the observation to a small subset of activated latent codes. We apply zero mean Gaussian noise with different variances on testing images to evaluate model's performance of denoising. In our case, we first try out three different trials with $\sigma=$ [0.3, 0.5, 0.7] using MNIST dataset. The latent variables are obtained from the inference step to reconstruct clean images from these noisy ones. 

\begin{figure}[ht]
\begin{center}
\centerline{\includegraphics[width=\columnwidth]{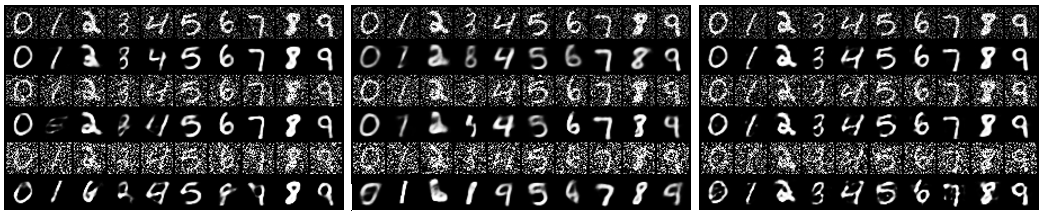}}
\caption{Denoising on MNIST Dataset. \textbf{Left:} Short-run inference model; \textbf{Middle:} VSC; \textbf{Right:} Our model. For each trial, odd rows represent noisy data obtained from zero mean Gaussian with standard deviation $\sigma$ = [0.3, 0.5, 0.7], even rows represent reconstructed images.}
\label{de1}
\end{center}
\end{figure}

In Figure \ref{de1}, we can observe that for short-run and VSC model, they will restore wrong digits when the noise variance is high. Their dense latent variables are sensitive to small changes in the latent space and they are not robust to the added noise. At the same time, our model can faithfully recover from the Gaussian noise with various noise level and obtain the correct digit. This is because the sparse model can learn key structure information of each class while pushing non-informative latent towards zero so they are not easily affected by the noise.

\begin{table}
\begin{center}

                    \begin{tabular}{llll}
                    \hline\noalign{\smallskip}
                    Model & $\sigma = 0.3$ & $\sigma = 0.5$ & $\sigma = 0.7$\\
                    \noalign{\smallskip}
                    \hline
                    \noalign{\smallskip}
                    VSC  & 0.33 & 0.17 & 0.11 \\
                    Short-run & 0.45 & 0.27 & 0.18\\
                    Ours  &  \textbf{0.48} & \textbf{0.33}  & \textbf{0.24}  \\
                    \hline
                    \end{tabular}

\end{center}
\caption{Performance on Structural Similarity Index (SSIM). Higher SSIM implies better image quality. Our model can produce best reconstructions from noisy images in terms of Luminance, Contrast and Structure.}
\label{SSIM}
\end{table}

In addition, we also calculate their Structural Similarity Index (SSIM) between noisy and denoised images to compare the perceptual changes and reconstructed image quality. Our model can outperforms both sparse  VSC model and dense short-run model in Table \ref{SSIM}. It implies the learned sparse latent representations using our algorithm are more robust to noises.

Moreover, we use training MNIST data to train a simple one layer classifier with ReLU activation to test the model's generalization on noisy images. By adding Gaussian noise with different variances to the testing data, we can obtain the latent representations from inference step and feed in the reconstructed images into the classifier to compute classification accuracy. The classification accuracy are averaged over 5 runs.

In Table \ref{class}, our model achieves the highest accuracy among other sparse and dense models. It shows our model is resistant to the perturbations from the Gaussian noise. The model has learned to encode important semantics into the sparse latent space while discarding latent codes with trivial contributions. So when there is noise added to the image, small perturbations will not have a large impact on the inference latent codes.

\begin{table}
\begin{center}

                    \begin{tabular}{llll}
                    \hline\noalign{\smallskip}
                    Model & $\sigma = 0.3$ & $\sigma = 0.5$ & $\sigma = 0.7$\\
                      \noalign{\smallskip}
                    \hline
                     \noalign{\smallskip}
                    VSC  & 72.41 & 57.94 & 53.54 \\
                    Beta-VAE & 72.31 & 62.13 & 59.49 \\
                    ABP & 74.16 & 47.74 & 55.39\\
                    Short-run & 78.78 & 74.23 & 68.17\\
                    Ours  &  \textbf{82.22} & \textbf{75.54}  & \textbf{71.52}  \\
                    \hline
                    \end{tabular}

\end{center}
\caption{Model classification accuracy on noisy images. Higher accuracy means the model can better capture structure information of the observations and robust to the added noise.}
\label{class}
\end{table}

\begin{figure}[ht]
  \begin{minipage}{0.50\linewidth}
    \centering
    \includegraphics[width=\linewidth]{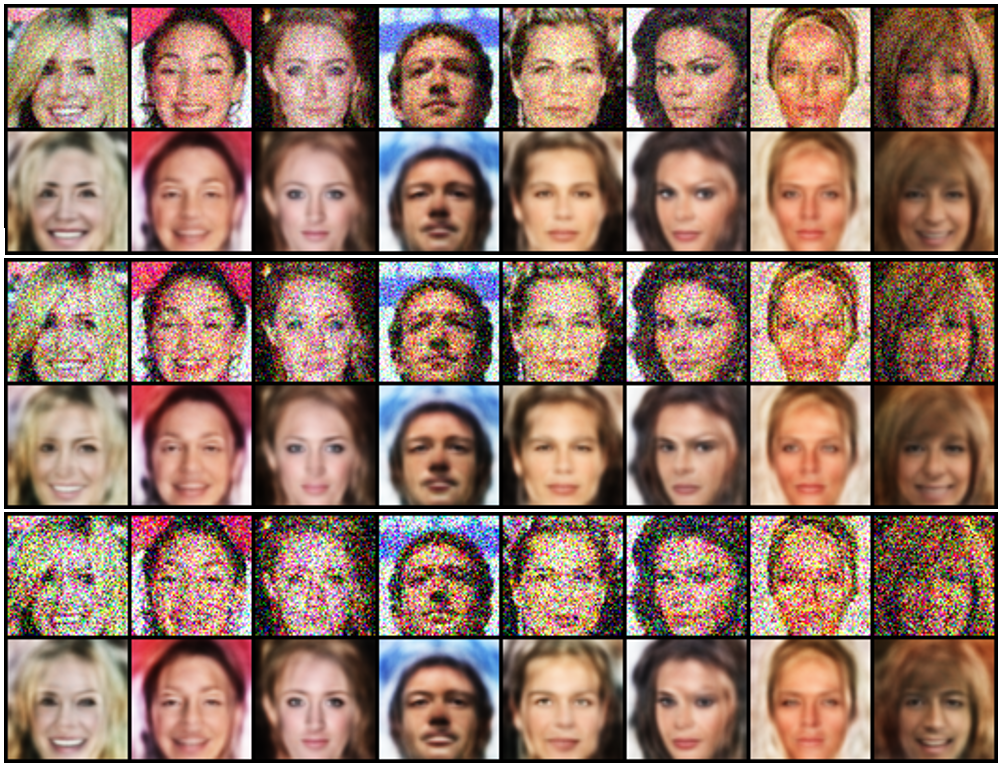}
    %\rule{6cm}{6cm} %to simulate an actual figure
  \end{minipage}%
  \begin{minipage}{0.50\linewidth}
    \centering
    \includegraphics[width=\linewidth]{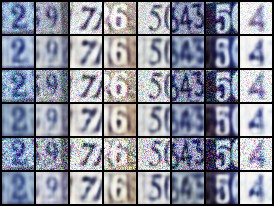}
    
    \end{minipage}

\caption{Denoising on CelebA and SVHN Images with various Gaussian noise. \textbf{Top rows:} Given noisy images. \textbf{Bottom rows:} Reconstructed Images.}
\label{denoise}
\end{figure}

For color images in Figure \ref{denoise}, we use $\sigma=$ [0.1, 0.2, 0.3] as the standard deviation for zero mean Gaussian noise. The noise is added to each image channel. We observe that our model can also reconstruct the corrupted face and digit images without changing too much of the observations. This implies the learned sparse latent representations can be robust to the Gaussian perturbation. The added distortion from noise does not have much impact on well-learned latent codes. The sparse representation is able to capture meaningful and structural information of the observed images.

%Even some of the figures are not visible to human, it can still correctly recover the digits from the noise compared to both ABP and VSC models.

\subsection{Ablation Study}

In this section, we investigate the effect of adding Langevin Dynamics diffusion term and the sparsity decay constant. We train these models using $\alpha = 0.01$ and 30 inference steps. We aim to evaluate the accuracy of reconstruction with learned latent codes and their performance on recovering structural knowledge from noisy images. We evaluate these models using PSNR and SSIM score. PSNR is obtained from reconstructed testing images and SSIM is calculated from noisy testing datasets using zero mean Gaussian noise with $\sigma = 0.1$ for color images and $\sigma = 0.5$ for grey scale images. Latent heatmap is also plotted with the same setting to visualize the learned structural information. \newline

\noindent \textbf{Langevin vs. gradient-based inference: } In Table \textcolor{red}{4}, we observe model with Langevin noise will lead to a smaller SSIM score and worse PSNR in majority of the dataset. The heatmap on the right side of Figure \textcolor{red}{10} also shows that the model is not able to capture meaningful structure of the given observations. The extra noise added during the inference step will distort the learned latent structure and make the latent representations less interpretable. The gradient-based inference can lead to more accurate latent representation. \\

\noindent \textbf{Sparsity decay vs. constant sparsity: } From Figure \textcolor{red}{10}, we observe that the model can learn some structural information from the images with constant sparsity. But it mixes some of the structural information into the same latent code for different category. Although this method can lead to better performance for certain datasets, its learned latent codes are not as disentangled as model with sparsity decay while some of them are always kept activated. A possible explanation to this phenomenon is that some of the latent codes are pushed towards zero at the start of the training when we have a small sparsity level. Starting with a dense model can give us the benefit to let each latent code has the ability to learn from images, and then non-informative codes will be forced towards zero as we gradually increase the sparsity value.

\begin{table}[t]
    
    \scalebox{0.65}{
  \begin{tabular}{lllllllll}
  \noalign{\smallskip}
    \hline
    \noalign{\smallskip}
    Model &
      \multicolumn{2}{c}{MNIST} &
      \multicolumn{2}{c}{Fashion} &
      \multicolumn{2}{c}{SVHN} &
      \multicolumn{2}{c}{CelebA} \\
    & PSNR & SSIM & PSNR & SSIM & PSNR & SSIM & PSNR & SSIM\\
    \hline
    Langevin Noise & 37.59 & 0.28 & 38.20 & 0.32 & 48.40 & 0.36 & 46.37 & 0.66 \\
    \hline

    Constant Sparsity& 39.09 & 0.33 & 38.71 & 0.37 & 50.75 & 0.41 & 45.03 & 0.67\\
    \hline 

    Proposed Model & 37.20 & 0.31 & 39.02 & 0.40 & 49.37 & 0.41 & 50.17& 0.69 \\

    \hline
    \label{ablation}

  \end{tabular}}
  \caption{PSNR and SSIM metric on different models.}
  \end{table}

\begin{figure}[H]
  \begin{minipage}{0.50\linewidth}
    \centering
    \includegraphics[width=\linewidth]{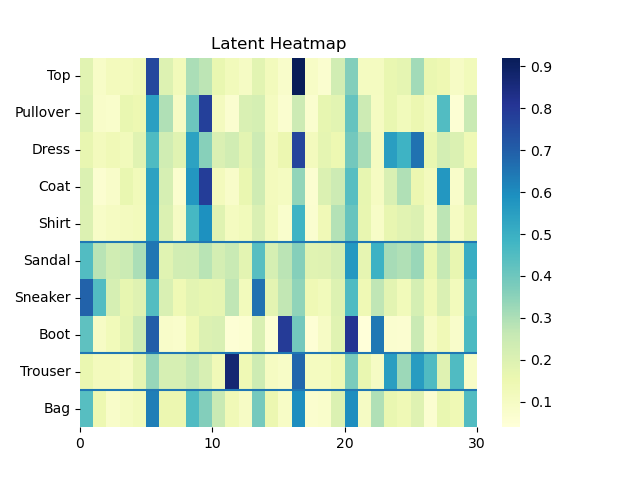}
    %\rule{6cm}{6cm} %to simulate an actual figure
  \end{minipage}%
  \begin{minipage}{0.50\linewidth}
    \centering
    \includegraphics[width=\linewidth]{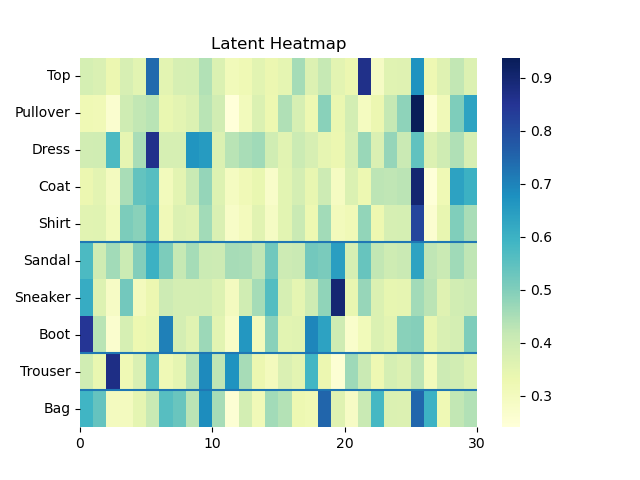}
    
    \end{minipage}

\label{ablation2}
\caption{Latent heatmap for ablation models. \textbf{Left: } Model without sparsity decay. \textbf{Right:} Model with Langevin Dynamics Noise.}
\end{figure}

\vskip -1in

\section{Conclusions}

In this work, we present a new learning method to learn a sparse latent representation with a gradually sparsified spike and slab as our prior distribution. The model uses only one top-down generator to map from the latent space to observed images. Latent variables are obtained from zero initialization and gradient-based inference with log-prior penalty. Our model shows competitive reconstruction capability with only few activated latent codes while preserving important information of the given observation. We also performed extensive experiments to demonstrate the sparse latent representation has improved explainability and boosted robustness in the task of denoising and latent classification.

\appendix
\section{Derivation of Equations}

\subsection{Log Gradient of Spike and Slab Regularization}
The spike and slab distribution can be viewed as:
\begin{align}
  p_{ss} (z) & = \alpha_1 N(0, \sigma_1^2) + \alpha_2 N(0,\sigma_2^2) \; \\
    & = \alpha_1 \frac{1}{\sqrt{2\pi}\sigma_1}e^{-\frac{z^2}{2\sigma_1^2}} + \alpha_2 \frac{1}{\sqrt{2\pi}\sigma_2}e^{-\frac{z^2}{2\sigma_2^2}}
\end{align}
Let $\alpha_2 = 1 - \alpha_1$ and take the log of both sides, we can obtain:
\begin{align}
    \log & p_{ss}(z) \\
    & = \log (\alpha_1 \frac{1}{\sqrt{2\pi}\sigma_1}e^{-\frac{z^2}{2\sigma_1^2}} + (1 - \alpha_1) \frac{1}{\sqrt{2\pi}\sigma_2}e^{-\frac{z^2}{2\sigma_2^2}}) \; \\
    & = \log (\frac{\alpha_1}{\sqrt{2\pi}\sigma_1}e^{-\frac{z^2}{2\sigma_1^2}}(1 + \frac{\frac{1-\alpha_1}{\sqrt{2\pi}\sigma_2}e^{-\frac{z^2}{2\sigma_2^2}} }{\frac{\alpha_1}{\sqrt{2\pi}\sigma_1} e^{-\frac{z^2}{2\sigma_1^2}}})) \; \\
    & = \log (\frac{\alpha_1}{\sqrt{2\pi}\sigma_1}) + \log (e^{-\frac{z^2}{2\sigma_1^2}}) + \log (1 + \frac{\frac{1-\alpha_1}{\sqrt{2\pi}\sigma_2}e^{-\frac{z^2}{2\sigma_2^2}} }{\frac{\alpha_1}{\sqrt{2\pi}\sigma_1} e^{-\frac{z^2}{2\sigma_1^2}}}) \; \\
    & = \log (\frac{\alpha_1}{\sqrt{2\pi}\sigma_1}) - \frac{z^2}{\sigma_1^2} + \log (1 + \frac{(1 - \alpha_1)\sigma_1}{\alpha_1\sigma_2}e^{\frac{(\sigma_2^2-\sigma_1^2)z^2}{2\sigma_1^2\sigma_2^2}})
\end{align}

\noindent We then take the partial derivative with respect to $z$ and get:
\begin{align}
    \frac{\partial}{\partial z} \log p_{ss}(z) & = 0 - \frac{2z}{2\sigma_1^2} + \frac{\frac{(1-\alpha_1)\sigma_1}{\alpha_1\sigma_2}e^{\frac{(\sigma_2^2-\sigma_1^2)z^2}{2\sigma_1^2\sigma_2^2}}\frac{2(\sigma_2^2-\sigma_1^2)z}{2\sigma_1^2\sigma_2^2}}{1 + \frac{(1-\alpha_1)\sigma_1}{\alpha_1\sigma_2}e^{\frac{(\sigma_2^2-\sigma_1^2)z^2}{2\sigma_1^2\sigma_2^2}}} 
\end{align}

\noindent Multiplying $e^{-\frac{(\sigma_2^2-\sigma_1^2)z^2}{2\sigma_1^2\sigma_2^2}}$ on the numerator and denominator of the third term, we can simplify the equation:

\begin{align}
    \frac{\partial}{\partial z} \log p_{ss}(z) & = -\frac{z}{\sigma_1^2} + \frac{\frac{(1-\alpha_1)\sigma_1}{\alpha_1\sigma_2}\frac{(\sigma_2^2-\sigma_1^2)z}{\sigma_1^2\sigma_2^2}}{e^{-\frac{(\sigma_2^2-\sigma_1^2)z^2}{2\sigma_1^2\sigma_2^2}}+\frac{(1-\alpha_1)\sigma_1}{\alpha_1\sigma_2}} \; \\
    & = -\frac{z}{\sigma_1^2} + \frac{R_1R_2z}{e^{-R_2\frac{z^2}{2}}+R_1}
\end{align}
where $R_1 = \frac{(1-\alpha_1) \sigma_1}{\alpha_1 \sigma_2}$ and $R_2 = \frac{\sigma_2^2 - \sigma_1^2}{\sigma_1^2 \sigma_2^2}$. With such derivation, we can prevent exponent overflow when training the model. 

\subsection{Posterior Component}
Given a Gaussian mixture model with two components and prior probability of component $p(C_i) = \alpha_i$, we can represent its posterior below:

\begin{align}
    p(C_i|z) 
    & = \frac{p(C_i)p(z|C_i)}{p_{ss}(z)} \; \\
    & = \frac{\alpha_i N(0, \sigma_i^2)}{\alpha_i N(0, \sigma_i^2) + \alpha_j N(0, \sigma_j^2)} \; \\
    & = \frac{\alpha_i \frac{1}{\sqrt{2\pi}\sigma_i}e^{-\frac{z^2}{2\sigma_i^2}}}{\alpha_i \frac{1}{\sqrt{2\pi}\sigma_i}e^{-\frac{z^2}{2\sigma_i^2}} + \alpha_j \frac{1}{\sqrt{2\pi}\sigma_j}e^{-\frac{z^2}{2\sigma_j^2}}} \; \\
    & = \frac{\frac{\alpha_i}{\sigma_i}}{\frac{\alpha_i}{\sigma_i} + \frac{\alpha_j}{\sigma_j}e^{-\frac{z^2}{2\sigma_j^2} + \frac{z^2}{\sigma_i^2}}} \; \\
    & = \frac{\frac{\alpha_i}{\sigma_i}}{\frac{\alpha_i}{\sigma_i} + \frac{\alpha_j}{\sigma_j}e^{z^2(\frac{1}{2\sigma_i^2}-\frac{1}{2\sigma_j^2})}}  \, \, \, i, j \in \{1, 2\}; i \neq j
\end{align}

\subsection{Log Gradient of Spike and Slab Prior from Posterior Perspective}

We start to take the derivative with respect to $z$ without any reparametrization from Equation 3:

\begin{align}
    \frac{\partial}{\partial z} &\log p_{ss}(z) \\
    & =  \frac{\partial}{\partial z}(\log (\alpha_1 N(0, \sigma_{1}^2) + \alpha_2 N(0, \sigma_{2}^2)))\\
    & =  \frac{\partial}{\partial z}(\log (\alpha_1 \frac{1}{\sqrt{2\pi}\sigma_1}e^{-\frac{z^2}{2\sigma_1^2}} + (1 - \alpha_1) \frac{1}{\sqrt{2\pi}\sigma_2}e^{-\frac{z^2}{2\sigma_2^2}})) \; \\
    & = \frac{-\alpha_1 \frac{1}{\sqrt{2\pi}\sigma_1}e^{-\frac{z^2}{2\sigma_1^2}\frac{2z}{2\sigma_1^2}} - (1 - \alpha_1) \frac{1}{\sqrt{2\pi}\sigma_2}e^{-\frac{z^2}{2\sigma_2^2}}\frac{2z}{2\sigma_2^2}}{\alpha_1 \frac{1}{\sqrt{2\pi}\sigma_1}e^{-\frac{z^2}{2\sigma_1^2}} + (1 - \alpha_1) \frac{1}{\sqrt{2\pi}\sigma_2}e^{-\frac{z^2}{2\sigma_2^2}}} \; \\
    \begin{split}
    & = -\frac{\alpha_1 \frac{1}{\sqrt{2\pi}\sigma_1}e^{-\frac{z^2}{2\sigma_1^2}\frac{z}{\sigma_1^2}}}{\alpha_1 \frac{1}{\sqrt{2\pi}\sigma_1}e^{-\frac{z^2}{2\sigma_1^2}} + (1 - \alpha_1) \frac{1}{\sqrt{2\pi}\sigma_2}e^{-\frac{z^2}{2\sigma_2^2}}} \\ & \qquad - \frac{(1 - \alpha_1) \frac{1}{\sqrt{2\pi}\sigma_2}e^{-\frac{z^2}{2\sigma_2^2}}\frac{z}{\sigma_2^2}}{\alpha_1 \frac{1}{\sqrt{2\pi}\sigma_1}e^{-\frac{z^2}{2\sigma_1^2}} + (1 - \alpha_1) \frac{1}{\sqrt{2\pi}\sigma_2}e^{-\frac{z^2}{2\sigma_2^2}}}
    \end{split} 
\end{align}
Substituting with Equation 14, we can have:
\begin{align}
    \frac{\partial}{\partial z} \log p_{ss}(z) = -p(C_1|z)\frac{z}{\sigma_1^2} -p(C_2|z)\frac{z}{\sigma_2^2}
\end{align}

%%%%%%%%% REFERENCES
{\small
\bibliographystyle{ieee_fullname}
\bibliography{egbib}
}

\end{document}